# Review on Multiple Plagiarism: A Performance Comparison Study

Jabir Al Nahian
Dept. of CSE
Daffodil International University
Dhaka, Bangladesh
jabir15-10414@diu.edu.bd

Abu Kaisar Mohammad Masum
Dept. of CSE
Daffodil International University
Dhaka, Bangladesh
abu.cse@diu.edu.bd

*Abstract*— Plagiarism is the practice of claiming to be someone else's content, thoughts or ideas as one's own without any proper credit and citations. This paper is a survey paper that, represent the some of the great research paper and its comparison that is work done on plagiarism. Now a days, plagiarism became one of the most interesting and crucial research points in Natural Language Processing (NLP) area. We review some old research paper based on different types of plagiarism detection and their models and algorithm, and comparison of the accuracy of those papers. There are many several ways which are available for plagiarism detection in different language. There are a few algorithms to detecting plagiarism. Like, corpus, CL-CNG, LSI, Levenshtein Distance etc. We analysis those papers, and learn that they used different types of algorithms for detecting plagiarism. After experiment those papers, we got that some of the algorithms give a better output and accuracy for detecting plagiarism. We are going to give a review on some papers about Plagiarism and will discuss about the pros and cons of their models. And we also show a propose method for plagiarism detection method which based on sentience separation, word separation and make sentence based on synonym and compare with any sources.

## I. INTRODUCTION

In this modern world, Internet is readily available for anyone and also anyone get any content which he wants.

For this reason, now a day's plagiarism is one of the most crucial problem in the world. In recent time that is mostly used in educational institution and research work. Most of the student do their assignment and educational task using others authors content which are easily get from internet and don't give proper credit and citations. Sometimes teachers cannot make their class materials by own. They also used others authors content for their class. Some of the researcher claim others authors research materials as own's research. Plagiarism is a deadly crime, but they did not feel it and they think that it is normal. For plagiarism, student cannot be giving their best effort for their educational task, assignment and others work, sometimes teachers cannot be able to evaluate their students for plagiarism, because of who copied internet content for their task and who do their task with his best effort. In the research field most of the time professional researcher cannot get their proper dignity for who copy research material and claim its own work.

As per an investigation on 18,000 understudies shows that about half of the understudies conceded they ordinarily appropriate their theses and tasks from superfluous records. As a result, it's critical that plagiarism detection systems can effectively detect this form of plagiarism. One of the greatest issues in writing and science is plagiarization. unauthorized utilize of the unique substance. Plagiarization is exceptionally troublesome to distinguish, particularly when the net is the source of data due to its measure. The location of plagiarization is indeed more difficult when is among reports composed totally different dialects. As of late a overview was done on researcher hones and demeanors [14].

A good plagiarism detector will help you save time and money. This paper discusses two methods for identifying plagiarism in documents from two separate sources [7]. The majority of plagiarism detection techniques, including the well-known Turnitin, don't really work in Bengali language. Despite the truth that Bangla is the world's seventh most commonly spoken language, this remains the case. Bangla scripts are freely accessible on a number of websites. Plagiarism is common in such textbooks, with the offenders gaining success and money instead of the actual creators. We want a successful and effective Bangla plagiarism detection tool to tackle Bengali language text fraud. This inspires us to develop a plagiarism detection system in Bengali language [1][3]. Understudies replicating programs frame others for a variety of purposes, including copying completed research, copying within the same source, coordinating on tasks (appropriate), reducing the amount of work required, and so on. Physically, it could be a strenuous and monotonous task, particularly when class sizes are large and hundreds of students are present.

A plagiaristic software is one that has been repeated with a small number of standard modifications [21]. Algorithm plagiarism detection may provide valuable insight into the identification of important algorithm characteristics. There hasn't been much research done on this subject before. Despite the fact that both algorithm and software plagiarism detection depend on comparing program similarities, they are radically different. They develop a free software framework for detecting plagiarism through languages [12]. Three similarity models were used to compare various forms of CL plagiarism. The sum of information, dialects, and time required make it inconceivable to perform in hone. Most of these procedures are planned for verbatim duplicates and execution is decreased when managing with light and especially overwhelming cases of copyright infringement, which incorporate paraphrasing [13].

In this paper, we overview and analyzing some paper we got some idea about plagiarism, its problem and plagiarism detection models and algorithms. Using those idea, we compare those model and algorithm and find best output for plagiarism and best accuracy for plagiarism detection result.

**Table.1** Abbreviation table

| Abbreviation | Algorithms/Model |
|---|---|
| CL-ASA | Cross-Language Alignment-based Similarity Analysis |
| CL-KCCA | Cross-Language Kernel Canonical Correlation Analysis |
| LSI | Latent Semantic Indexing |
| LCCS | Least Congested Channel Search |
| CPBD | Citation-based Plagiarism Detection |
| TF-IDF | Term frequency-inverse document frequency |
| CL-CTS | Cross-Language Conceptual Thesaurus-based Similarity |
| CL-ESA | Cross-Language Explicit Semantic Analysis |

## II. OVERVIEW

In this section, we give an overview-based plagiarism detection paper. Plagiarism detection tool is alike a traditional machine-readable tool which compare to composition and give a result. Its plagiarized or not. It takes input as a text and compare with many sources and then show result. Using corpus creation method, take any text as an input. Then tokenize the text. After tokenize remove stop words and counting TF-IDF scores of each model. At last, compare two text (that means original text and suspicious text) using cosine similarity algorithm and show the result [1][3]. They investigated the method's complexity and performed detailed tests on a series of real records to demonstrate the efficacy and reproducibility of our ideas. Second, they explained the strategy for identifying root word that improves the productivity of literary Plagiarism identification. Next, they gave a procedure to identifying literary plagiarism dependent on synonyms and keywords in the records. Fourth, we conducted a quantifiable analysis of the documents and, based on the statistical results, they identified copyright infringement in the archives [7]. We read another paper which was written about cross language and they used The LSI concept is used to create a CLS space on which it compares the logic of two thesis articles, one was in Arabic and the another in English. Latent Semantic Indexing is a well-known data recovery technique. It assesses comparability on a contextual rather than linguistic basis. Actually, Words and reports are projected in a three-dimensional space by LSI [4][8].

We discovered that cross-language plagiarism, as depicted in another article, refers to circumstances in which a writer interprets text from another dialect and then integrates the resulting interpretation into his or her own composition. We go through the CL-ESA model, the model CL-ASA, as well as the CL-CNG model, which are all used to evaluate cross-language writing comparability. Both of the experiments were rerun using test sets culled from the JRC-Acquis corpus and the Wikipedia document equivalent. Customized reports in German, Polish, German, English, French, and Spanish are included with each test collection [5][12]. Despite the fact that plagiarism is the most well research area with dozens of stories, no journals or publicly available systems have considered using citation information to detect plagiarism. Due to a lack of research into the use of citation data, the researcher proposes a citation pattern analysis method for plagiarism detection, which he called Citation-based Plagiarism Identification method that use citations and references to recognize similarities between documents in order to detect plagiarism is recognized as citation-based plagiarism detection. [8]. In arrange to restrain certain scale of our proposed framework, we are considered Bahasa Melayu as an input dialect of the submitted inquiry record and English as a target dialect of comparable, conceivably copied documents.

We coordinated the utilize of Stanford Parser and WordNet to decide the closeness level between the suspected archives with those candidate source documents. Input archives are deciphered into English utilizing Google Decipher API some time recently experience pre-processing stage (stemming and expulsion of halt words). As it were best ten sources recovered by the Google Look API are considered as the candidate of source reports [15]. They show an approach to producing such rundowns, a covered-up Markov show that judges the probability that each sentence ought to be contained within the rundown [16].

These algorithms extricate a few program metrics highlights from a program and utilize this set of measures or highlights to compare programs for copyright infringement. The whole prepare can be seen as a classical design acknowledgment framework which extricates a number of features from the input information and after that employments factual remove or relationship measures on this include set to compare the designs or programs. Since the usage of these calculations requires a compiler-like front conclusion, an adjusted form of the calculation is required for each programming dialect. The extraction of these highlights requires checking, looking a saved word table, development of an image table, and restricted parsing of the programming language [18].

After analyzing the all papers, we found some difficulty and limitation of those paper. Like as, they didn't study about related topics. They don't provide some information; those are need in papers. Example: How can they identify paraphrased sentence, how can they handle their algorithm etc. Some of the papers we see that they are highly focused on specific site of plagiarism. Now we give out some case, some authors focus only English language plagiarism, some are focus only Arabic and some are focus only Bangla language plagiarism etc.

Indeed, our main focus is building a plagiarism technique which is applied for not only specific language but also multiple language plagiarism.

**Table.2** Plagiarism detection comparability table

| Article | Subject | Observed Features | Functionality | Models/Algorithms | Results |
|---|---|---|---|---|---|
| 01 | Bangla Language | Established a corpus that includes all books from Bangladesh's Public Educational Program and Course Reading Board (NCTB) from classes I to XII. Migrate lesson books into plain, writable Bangla texts. Project in docx format. At that point, Tokenization comes into play (split removed writings) | Detects similarity between Bangla texts. | Corpus Creation Pre-Vectorization TF-IDF algorithm Cosine and Jaccard similarity algorithm | 96.75% Accuracy |
| 02 | Paper Plagiarism | Measures whether the contiguity between plagiarized documents entries is appropriately recognized. | Detects similarity various research paper. | Corpus Creation | 87.63% Accuracy |
| 03 | Paper Plagiarism | Executes a plagiarism search space decrease strategy, and afterward executes a comprehensive inquiry to discover copied sections. | Intrinsic plagiarism detection seeks to identify plagiarism by analyzing only the input text and determining if sections of it are not by the same author. External plagiarism recognition is a method that compares suspect documents to a list of possible references. A frequency-based algorithm for determining document self-similarity. | Intrinsic plagiarism evaluation External plagiarism evaluation corpus PAN2009 corpus PAN2011 | Intrinsic plagiarism: 34% External plagiarism: 80% |
| 04 | Arabic and English Language | Compares comes about of detecting plagiarizing between sentence comparison and semantic comparison | Contextual similarity of two given research papers. | Latent Semantic Indexing (LSI) | 93% Accuracy |
| 05 | Dutch, French, Spanish, German, English, and Polish | In this observation, the closeness proportions of comparable and similar documents are compared. | A detailed cross-language analysis of two documents | CL-CNG CL-VSM CL-ASA CL-KCCA | Wikipedia: 55% JRC-Acquis: 66% |
| 06 | English Language | In the field of text comparison, investigate relevant heuristics. | Examine the application of a diagonal line. | Levenshtein Distance and Smith-Waterman Algorithm | Levenshtein Distance: 94.44% Smith-Waterman Algorithm: 95.52% |
| 07 | English and Bangla Language | The first approach involves performing various statistical analyses of the documents in order to identify plagiarism, while the second method is focused on analyzing the individual components of the documents. | The first method entails conducting various statistical analyses of the documents in order to detect plagiarism, while the second method focuses on evaluating the documents' individual components. | Keywords Extraction | Bangla Language 57% Accuracy English Language 73.20% Accuracy |
| 08 | Citation-based | Find lexical and semantic similarity between documents. | To detect plagiarism, look for similarities between documents. | LSI LCCS GCT Citation Chunking | 34% Accuracy |
| 09 | English and Spanish | Covering the whole prepare: heuristic recovery, point by point examination, and post-processing. The three models are tried broadly beneath the same conditions on the distinctive copyright infringement discovery sub-tasks—something never done some time recently. | Plagiarism detection in English language and also cross checking with Spanish language. | CL-ASA CL-CNG CLPD | JRC 80% Accuracy INF 60% Accuracy STEM 78% Accuracy |
| 10 | English, Spanish and German Language | Content sections that use knowledge charts as a dialect-independent substance display. We investigate the commitments to cross-language copyright infringement position of information graphs' various viewpoints: word - based lexicon extension, disambiguation and representation by similitudes with a set of ideas. | The results of experiments in plagiarism detection between Spanish and English also German and English display province success as well as provide important perspectives into the use of information graphs. | CL-KGA CL-ESA CL-CNG | 68% Accuracy |

## III. METHODOLOGY

Tasks of Learning: At first, we take two texts, after that taken text, we divided into two division. One division for learning model and another is checking plagiarism. We split every sentence in too composition and dubious text. Since split, we extracted the stop words from every sentence. We also follow the instruction of first paper. They propagate TFIDF [1] They compare these values using the Cosine Similarity algorithm to generate similarity between two texts, scoring each sentence in both composition and dubious text. If the equality exceeds a certain point, they admit that a portion of the text has been plagiarized. It also highlights plagiarized sentences and the key composition's footprint. The framework shows a total report of literary theft of the dubious record [1][3]. We also choose another model which based on LSI. This paper broadly works by vocabularies and its translations for plagiarism detection. LSI has proven to be an excellent tool for cross language rehabilitation. Commonly, the interaction depends on a teaching and learning process of multi-language documents as opposed to utilizing direct interpretation. For our situation, we address the vocabularies of both language in Latent Semantic Indexing location. Vocabularies that are reliably matched or every now and again connected with each other. Setting a record in the LSI field now has a language representation, whether in English or Arabic. Anyone, regardless of language, can search for plagiarism in double language and get similar documents [4].

### A. Pre-Vectorization and Corpus Creation

Corpus is a grouping of composed compositions and most critical component of our plagiarization discovery device. We built up corpus on the educator region by downloading each one of the course readings of Lesson I-XII NCTB aside from English composing books and can be found in. Each one of these books requires planning some time recently at final included to corpus. Preprocessing incorporates the taking after steps:

1. Extraction of Text: Text Extraction is a cycle by which we convert Printed report/Scanned Page or Image in which text are accessible to ASCII Character that a computer can Recognize. The reading material are accessible in Portable Document format only. To start with, we change over each one of the reading materials into writable doctor and document format. After that, we remove all of the tables and pictures from the books, as well as all of the extra whitespace and tabs. This results in a book with only plain Bangla content material. An entire content extraction process is performed physically, and it takes a long time. In this vein, we're developing self-contained applications to complete this task.

2. Tokenization: We know that, Tokenization is a method of separating text based on a format string as an example tabs, characters, punctuations, newlines, and so on. "(Bangla full stop)" is used to distinguish messages here. This creates sentences that are distinct, which we then cut any unnecessary whitespace. The tokenization is handled by the Python NLTK library's capabilities.

3. Stop words Removal: The stop word in a sentence is one or more words that contains no important detail. As a result, stop words are sifted through after tokenization and before normal language data is prepared. By integrating and subtracting, repetitive prevent words via various sources, we are capable of creating Bangla stop word datasets.

4. Scores Generation of TF-IDF: Following that, remove stop word. Then apply TFIDF Algorithm for our result. First, calculate TF value from text and again calculate TF result from sentence. Next, compute TF and IDF score. Finally, we stock TFIDF result in corpus.

5. Plagiarism Detection: After TF-IDF process using plagiarism detection algorithm. First, we compute TF and IDF score from suspicious text. Then put every corpus sentence in corpus database. Fetch TFIDF result of corpus sentence from corpus. Now, Compute cosine plagiarism. If the similarity is greater than suspicious sentence then we call it plagiarized [1].

### B. Latent Semantic Indexing (LSI)

Latent Semantic Indexing is a hypothesis and strategy for separating and addressing the logical utilization significance of words by measurable calculations. It isn't straightforward contiguity frequencies or co-event, however relies upon a more profound a comparative review. It is valuable in circumstances where classical lexical data recovery techniques are unsuccessful. LSI gauges the semantic substance of the archives in an assortment and utilizations this gauge to rank the records arranged by diminishing pertinence to a client's question. Latent Semantic Indexing (LSI) is valuable in circumstances, where search depends on the ideas contained in the reports as opposed to the record's constituent terms, LSI can recover documents identified with a client's inquiry in any event, when the question and the reports don't share any normal terms. It performs well in the old-style issues of synonymy and polysemy. LSI utilizes same method of human judgment on likeness. Various exploratory web crawlers were created utilizing LSI methods. LSI has been utilized in cross-language recovery consummately. Ordinarily, the interaction depends on a preparation cycle of double language reports as opposed to utilizing direct interpretation [4][8].

### C. Levenshtein Distance Algorithm

A widely utilized based up potential programming algorithm for manipulating the Levenshtein distance is with a $(m + 1)$ $(n + 1)$ matrix, since m and n are the ranges of the two sentences. The Wagner-Fischer edit distance algorithm is used to compute this value. [6].

### D. Keywords Extraction

Using a morphological analyzer, the keyword extractor sub-module extracts the base of each expression. The keyword processor's equation accepts a word as input and returns the yield of the terms. Stemming is the process of determining a word's origin. [7]

*E. Parallel corpus-based systems*

Parallel corpora are used to train these structures, both to identify cross-language professional and non or to acquire interpretation modules. Machine conversion concepts and tools are used, but no direct translating is done [12].

*F. Similarity Analysis Using Graphs*

To use a sentence distance matrix on the input text document, Fragment the entire text into a series of paragraphs. The sections are labeled with their lexical system using the infinitive form of the words. Using Tree Tagger 3, which allows several translations, for our tests. The tagged subsections' information graphs are generated. An information graph is a measured and classified graph that represents an article's concepts. And MSN peer definitions including marked connections based on their relationships. Extending the initial language by using information graphs to build a meaning model from the input document [14]. Compare those documents and show similarity by compare these graphs.

## IV. PROPOSED METHOD

We analyzing some paper about plagiarism detection, we get some good ideas about plagiarism detection and cross language checking. We think a method, to detect plagiarism problem. In fig. 1 we present our proposed method. We will collect data from school textbook, College textbook, newspaper, different types of articles and different types of blogs etc. Here, a user gives an input composition or text in the system. After taking input from user system separate each sentence from input composition or text. After separation all sentence from the text then the system automatically separates each word from every sentence. After that the system find and collect of every word's synonyms from synonym database, which word are in input composition or text.

After collecting text database, we create synonyms database using dictionaries. Then the system makes different type of sentence based on main sentence using all synonyms. After that, System check those sentences compare with all source database. After compare the whole input composition or text with sources if show result. If the composition matched with any sources it shows its plagiarized document, else it's an authentic document.

In our proposed method best thing is, it works not only for documents but also words plagiarism detection. Out method also detect and showing text plagiarism percentage and also word plagiarism percentage too. We see that, many different language plagiarism detections already used in different language like English, Spanish, China etc. But we cannot see such any good work about Bangla Language plagiarism detection and multiple plagiarism detection. So, we will try to our best to give a good efficiency in Bangla plagiarism detection also multiple language plagiarism detection. So that, if we can implement our propose method in real life, we thing it gives better feedback in multiple language plagiarism technique.

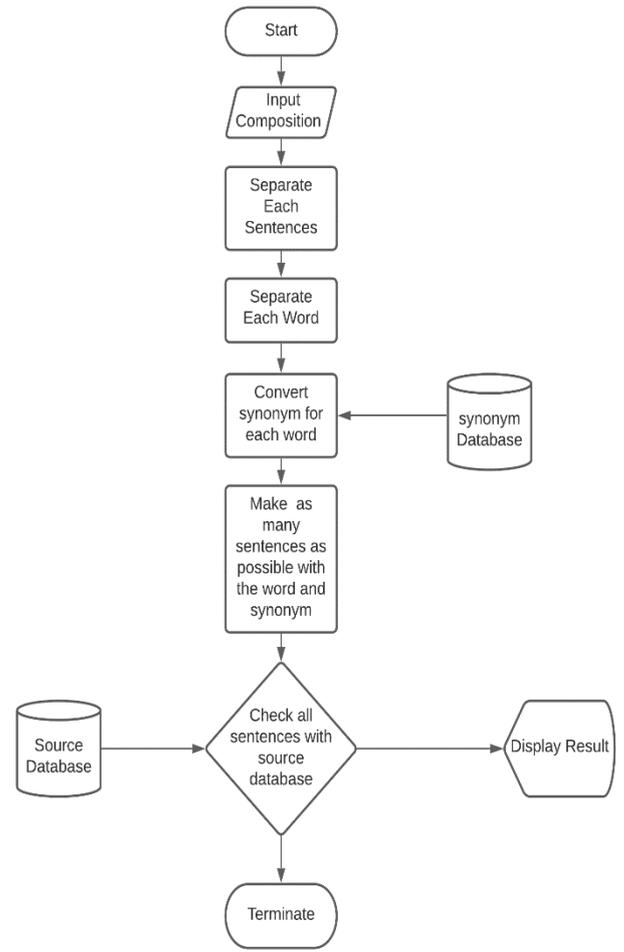

Fig.1 Plagiarism Detection Model

## V. DISCUSSION AND CONCLUSION

In this modern era, Internet is very easy to use for anyone. For this reason, plagiarism tendency increasing day by day. Not only student but also teacher, researcher and most of them are include with plagiarism.

Many plagiarism detection tools are already invented. Like as, Same language based, translation based etc. But we cannot find any plagiarism detector for Bangla language. Even, most of the plagiarism detection tool do not work properly. We analyzing some paper about plagiarism detection, we get some good work. We read a paper which is framework based and its working on paper plagiarism. They use Corpus Creation algorithm and get 87.63% Accuracy [2]. Also one another paper working on English and Arabic language. Using Latent Semantic Indexing (LSI) And Its output very appreciable and Its 93% Accuracy [4]. We found a paper, where Author collect data from textbook. And they make Deep Neural Networks approach. They apply to their approach Cosine and Jaccard similarity algorithm and get 96.75%. They have successfully introduced a plagiarism detection method depend on close domain for the Bengali language [1]. Although, we finish study more and more about plagiarism detection technique, but we have some limitations. First limitation, we cannot find our valuable dataset for our model. So, we can not implement our propose method properly. As things currently stand now, we collect data for making a proper dataset for our propose method. Every study intended to increase the efficiency of the plagiarism detection system by using a Natural Language Processing approach. In future, we want to work with Open Domain Bangla Plagiarism Detection tool and multiple language plagiarism detection.

## ACKNOWLEDGMENT

We would like to express our gratitude to our Computational Intelligence LAB for providing all research resources. Our supervisor and course instructor deserve special thanks for his patience and support in helping us overcome the lot of obstacles we encountered during our research.